\documentclass{article}

     \PassOptionsToPackage{numbers, compress}{natbib}

\usepackage[final]{neurips_2022}

\usepackage[utf8]{inputenc} 
\usepackage[T1]{fontenc}    
\usepackage{hyperref}       
\usepackage{url}            
\usepackage{booktabs}       
\usepackage{amsfonts}       
\usepackage{nicefrac}       
\usepackage{microtype}      
\usepackage{xcolor}         

\usepackage{amsmath}
\usepackage{graphicx}
\usepackage{subfigure}
\usepackage{algorithm}
\usepackage{algorithmic}
\usepackage{stmaryrd}
\usepackage{multicol,lipsum}
\usepackage{caption}

\renewcommand{\vec}[1]{\ensuremath{\mathbf{#1}}}

\newcommand{\mat}[1]{\ensuremath{\mathbf{#1}}}

\newcommand{\ten}[1]{\mat{\ensuremath{\boldsymbol{\mathcal{#1}}}}}

\newcommand{\norm}[1]{\left\lVert#1\right\rVert}
\title{SHINE: SubHypergraph Inductive Neural nEtwork}

\author{%
  Yuan~Luo
    \\
  Feinberg School of Medicine\\
  Northwestern University\\
  Chicago, IL 60611 \\
  \texttt{yuan.luo@northwestern.edu} \\
}

\begin{document}

\maketitle

\begin{abstract}
Hypergraph neural networks can model multi-way connections among nodes of the graphs, which are common in real-world applications such as genetic medicine. In particular, genetic pathways or gene sets encode molecular functions driven by multiple genes, naturally represented as hyperedges. Thus, hypergraph-guided embedding can capture functional relations in learned representations. 
Existing hypergraph neural network models often focus on node-level or graph-level inference. There is an unmet need in learning powerful representations of subgraphs of hypergraphs in real-world applications. For example, a cancer patient can be viewed as a subgraph of genes harboring mutations in the patient, while all the genes are connected by hyperedges that correspond to pathways representing specific molecular functions. 
For accurate inductive subgraph prediction, we propose SubHypergraph Inductive Neural nEtwork (SHINE). SHINE uses informative genetic pathways that encode molecular functions as hyperedges to connect genes as nodes. SHINE jointly optimizes the objectives of end-to-end subgraph classification and hypergraph nodes' similarity regularization. SHINE simultaneously learns representations for both genes and pathways using strongly dual attention message passing. The learned representations are aggregated via a subgraph attention layer and used to train a multilayer perceptron for inductive subgraph inferencing. We evaluated SHINE against a wide array of state-of-the-art (hyper)graph neural networks, XGBoost, NMF and polygenic risk score models, using large scale NGS and curated datasets. SHINE outperformed all comparison models significantly, and yielded interpretable disease models with functional insights. 
\end{abstract}

\section{Introduction}

Hypergraph neural networks have recently emerged as a series of successful methods to model multi-way connections that are beyond pairwise associations among nodes of the graphs. Multi-way connections are common in many real-world applications and, in particular, genetic medicine. From genetic medicine's perspective, pathways or broadly speaking gene sets encode the relationship among multiple genes that collectively correspond to a molecular function~\citep{liberzon2015molecular}, which can be used in machine learning models to account for disease mechanisms more intuitively and accurately than individual genes~\citep{luo2021panther}. Genetic pathways or gene sets encode functional relations among multiple genes (see Appendix for detailed explanations), which can be naturally modeled as hyperedges connecting all involved nodes (e.g., genes). Thus, hypergraph-guided embedding can capture functional relations in learned representations.

Existing hypergraph neural network models often adopt the semi-supervised learning (SSL) paradigm to assign labels to initially unlabeled nodes in a hypergraph~\citep{feng2019hypergraph,yadati2018hypergcn,yadati2020neural}. Other methods have focused on learning graph representations~\citep{zhang2019hyper,ding2020more}. Node-level and graph-level representations give either local or overarching views of the graphs, i.e., at the two extremes of hypergraph topological structures. There is an unmet need in learning powerful representations of subgraphs in hypergraphs. Such capabilities are important in real-world applications such as genetic medicine. For example, cancer patients can be viewed as subgraphs of genes that harbor mutations, while all the genes are connected by hyperedges that correspond to pathways or gene sets representing specific molecular functions. Powerful subgraph representations will lead to the capability to more accurately account for the patient's pathophysiology. For regular graphs where edges connect node pairs, several subgraph representation learning algorithms were proposed, including methods that can use the learned representations to make predictions for subgraphs with fixed sizes~\citep{meng2018subgraph} or varying sizes~\citep{alsentzer2020subgraph}. There are currently few if any work on inductive inference for varying-sized subhypergraphs. In this work, we propose a new framework named SHINE: SubHypergraph Inductive Neural nEtwork. We share our source code at https://github.com/luoyuanlab/SHINE. Our contributions are as follows:

\begin{itemize}
\setlength\itemsep{0em}
\item To the best of our knowledge, SHINE is the first model to effectively learn subgraph representations for hypergraphs, use the learned representations (for seen subgraphs) and inductively infer representations (for unseen subgraphs) for downstream subgraph predictions.
\item Novel applications in the field of genetic medicine on Next Generation Sequencing (NGS) datasets across diverse diseases show significant performance improvements by SHINE over a wide array of state-of-the-art baselines. 
\item In addition to learning and inductively inferring subgraph representations, SHINE simultaneously learns the representations of nodes and hyperedges. This brings interpretation advantages, allowing assessing pathways (hyperedges) correlations  and reasoning on multiple molecular functions mutually interacting and collectively contributing to disease onset and progression.
\end{itemize}

\section{Related Work}
\textbf{Graph Neural Networks.} Graph representation learning maps graphs or their components to vector representations and has attracted growing attention over the past decade. Recently, graph neural networks (GNNs), which can learn a distributed representation for a graph or a node in a graph, are widely applied to a variety of areas including computer vision and image processing~\citep{wei2020view,mao2022imagegcn}, molecular structure inference~\citep{duvenaud2015convolutional,gilmer2017neural}, natural language processing~\citep{yao2019graph,peng2018large,li2019classifying}, and healthcare~\citep{zitnik2018modeling,mao2019medgcn}. GNN recursively updates the representation of a node in a graph by aggregating the feature vectors of its neighbors and itself, e.g.~\citep{kipf2016semi}. The graph-level representations can then be obtained through set pooling (e.g.,~\citep{vinyals2015order}) or graph coarsening (e.g.,~\citep{ying2018hierarchical}) to aggregate the node representations in the graph. The reader is referred to a comprehensive book~\citep{hamilton2020graph} on the topic of graph neural networks.

\textbf{Hypergraph neural network.} Hypergraph neural networks~\citep{zhang2019hyper,yadati2018hypergcn,feng2019hypergraph,ding2020more} have become a popular approach for learning on multi-way relations from data. Early work on hypergraph learning, e.g., ~\citep{zhou2006learning}, formulated hypergraph message passing using spectral theory of hypergraphs. This formulation and its variants~\citep{feng2019hypergraph,yadati2018hypergcn,jin2019hypergraph,jiang2019dynamic,satchidanand2015extended,feng2018learning} essentially adopted clique expansion to extend graph convolutional network (GCN) for hypergraph learning. Others methods applied attention mechanism to aggregate the information across the hypergraph~\citep{zhang2019hyper,ding2020more} or directly learned node representations to preserve the proximity of nodes sharing a hyperedge or having similar neighborhoods~\citep{tu2018structural}. In both formulations, messages were passed to the node of interest from its immediate neighbors, and added layers allow propagation of messages to a farther neighborhood. A very recent model implements hypergraph neural network layers in a generalized way as compositions of two multiset functions that are approximated by neural networks~\citep{chien2021you}.   

\textbf{Subgraph representation learning and prediction.} Recent studies on subgraph embeddings and prediction starts with learning representations of small subgraphs. ~\citep{meng2018subgraph} encoded small fixed-sized subgraphs for subgraph evolution prediction. SubGNN~\citep{alsentzer2020subgraph} learned representations for varying-sized subgraphs through neighborhood, position and structure channels using random patches distributed throughout the graph. ~\citep{huang2020graph,sun2021sugar} learned subgraph representations by pooling local structures to aid the predictions of the entire graphs.  Note modeling hyperedges as another type of nodes turns hypergraphs into bipartite (and heterogeneous) graphs,  making SubGNN a potential baseline for subhypergraph inferencing. On the other hand, most of the existing general heterogeneous graph neural network models do not support subgraph inferencing~\citep{wang2019heterogeneous,zhang2019heterogeneous,fu2019metapath,fu2020magnn,hu2020heterogeneous}. 

The intersection of hypergraph neural network and subgraph representation learning is currently underexplored. While the above methods focus on either hypergraph learning or subgraph learning, none of the methods consider subgraph prediction for hypergraphs.  Technically, subgraphs can be viewed as a hyperedge and studies on link prediction could predict the existence of a hyperedge~\citep{zhang2018beyond}. However, few if any such studies addressed the problems of differentiating the classes of the subgraphs, which is especially important in genetic medicine where subgraphs and hyperedges have different real-world meanings. For example, a hyperedge corresponds to a genetic pathway from curated knowledge and a subgraph corresponds to a patient with mutated genes as its nodes.

To sum, there is a major unmet need regarding varying-sized subgraph inference for hypergraphs, and even more so in the inductive learning setting. Our proposed framework SHINE provides an end-to-end framework that operates on hypergraphs and performs inductive subgraph inferencing. 

\section{Methods}

We first outline the workflow of SHINE, see Table~\ref{tab:notation} for symbol definitions. We develop a strongly dual attention message passing algorithm to propagate information between nodes and hyperedges, and across layers. We develop a weighted subgraph attention mechanism to learn the subgraph representation by integrating representations of hypergraph nodes. We next explain each step.

\begin{table}
\centering
\footnotesize
\caption{Common notations used throughout the paper.}
\begin{tabular}{lr|lr}  
\toprule
Symbol  & Definition & Symbol  & Definition \\
\midrule
$\ten{H}$ & An undirected hypergraph & $\ten{N}$, $|\ten{N}|$ & Set of hypergraph nodes and size \\
$\ten{E}$, $|\ten{E}|$ & Set of hypergraph hyperedges and size & $d$ & Hidden layer size \\
$\mat{H}$ & Hyperedge incidence matrix & $\circ$ & Operation composition \\
$n$ & Number of subgraphs & * & Element-wise multiplication \\

\bottomrule
\end{tabular}
\label{tab:notation}
\end{table}

\subsection{Collecting Genetic Pathways}
We use the Molecular Signatures Database (MSigDB)~\citep{liberzon2015molecular} and focus on MSigDB's curated pathway (gene set) collection, which contains human gene sets that are canonical representations of a biological process compiled by domain experts.  There are 21,587 genes in MSigDB Pathways. Some pathways may overlap with others and have been filtered by MSigDB to remove interset redundancy. Genes with unknown functions are not included in the pathways and not used for classification, as our focus here is on interpretable modeling through inferencing with known molecular functions. Adding genes with unknown functions to study their impact will be our future work.

\subsection{Hypergraph Learning}
We first review the hypergraph analysis basics. Different from a simple graph, a hyperedge in a hypergraph connects
two or more vertices. A hypergraph is defined as $\mathcal{H} = (\mathcal{N}, \mathcal{E})$, which includes a set of nodes $\mathcal{N}$, a set of hyperedges $\mathcal{E}$. In the case of genetic medicine, we model genes as hypergraph nodes, i.e., $\mathcal{N}=\{g_1, ...,g_{|\mathcal{N}|}\}$, and pathways as hyperedges, i.e., $\mathcal{E}=\{p_1, ...,p_{|\mathcal{E}|}\}$, where $|\mathcal{N}|, |\mathcal{E}|$ are sizes of the nodes and hyperedges respectively. The hypergraph's topological structure can be denoted by an $|\mathcal{N}| \times |\mathcal{E}|$ incidence matrix $\mat{H}$, whose entries are defined as 
\begin{align}
\mat{H}_{ij} = 
\begin{cases}
\quad 1 & \text{if node } g_i \in \text{ hyperedge } p_j \\ 
\quad 0 & \text{if node } g_i \notin \text{ hyperedge } p_j\\
\end{cases}
\label{eq:graph-str}
\end{align}
Generally speaking, each node in the hypergraph may be accompanied by a $d$-dimensional node feature/embedding matrix $\mat{N} \in R^{|\mathcal{N}| \times d}$, where each row corresponds to a node's feature/embedding. The hypergraph with its topological structure and node features can be represented succinctly as $\mathcal{H} = (\mat{H}, \mat{N})$. 

\begin{figure*}[t]
    \centering
    \includegraphics[width=\textwidth]{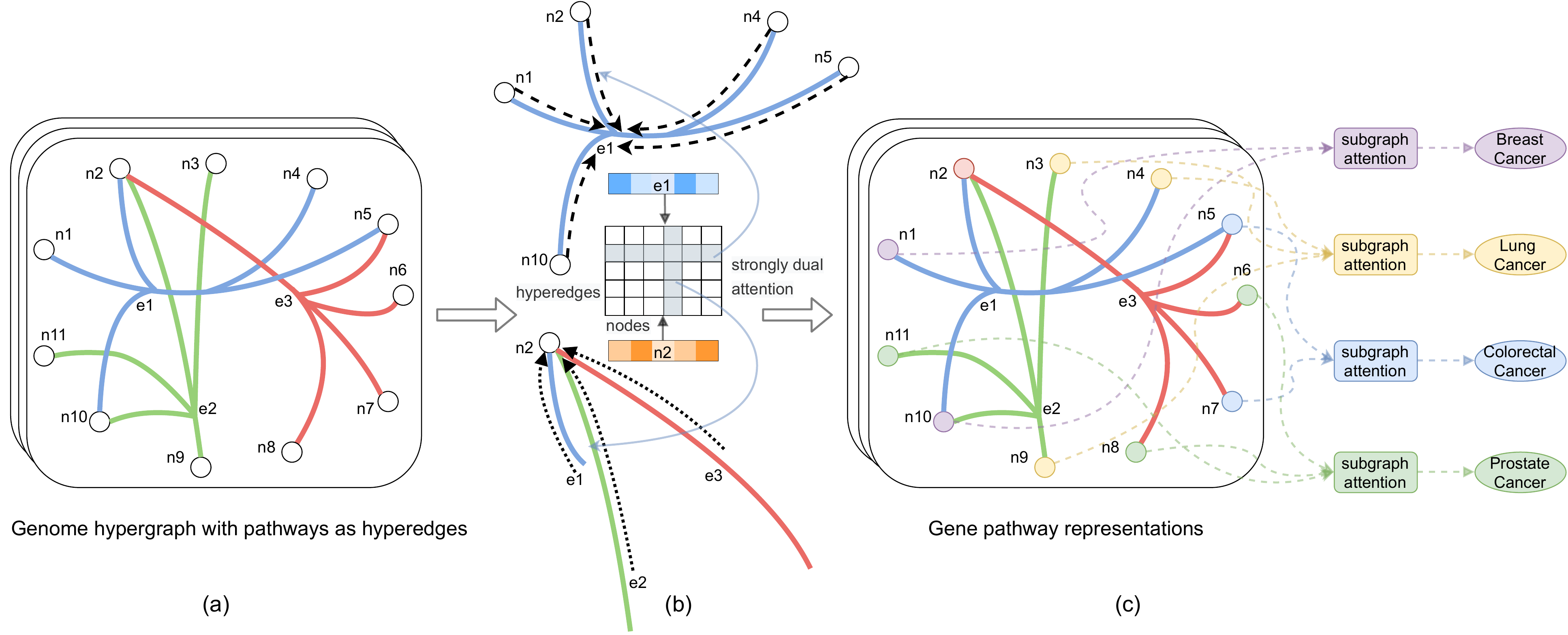}
    \caption{SHINE's strongly dual attention mechanism for message passing for the genome hypergraph, and its use of subgraph attention to integrate gene nodes in the feature learning for subgraphs.}
    \label{fig:SHINE}
\end{figure*}

Fig.~\ref{fig:SHINE} (a) shows a schematic of the constructed genome hypergraph with nodes denoted by circles and hyperedges denoted by colored arcs. While a pathway can contain multiple genes, a gene can also contribute to multiple pathways. That is, we can have multiple hyperedges incident on the same node (gene), as can be seen in Fig.~\ref{fig:SHINE} (a) nodes $n_2, n_5, n_{10}$. 

\subsection{Strongly Dual Attention Message Passing}
\textbf{Hyperedge attention over nodes.} For a hyperedge $p_j \in \mathcal{E}$, in order to update its hidden representation at layer $k$, we aggregate the information from its incident nodes using the following attention mechanism. We first calculate the hyperedge attention over nodes as in
\begin{equation}
a_E(p_j,g_i) = \text{exp}\left( \vec{c}^T \vec{s}(p_j,g_i) \right) \bigg/ \left(\textstyle\sum\nolimits_{g_{i'} \in p_j} \text{exp}\left( \vec{c}^T \vec{s}(p_j,g_{i'}) \right) \right)
\end{equation}
where $\vec{c}$ is a trainable context vector and  the attention ready state $\vec{s}(p_j,g_i)$ for a hyperedge-node pair $(p_j,g_i)$ is calculated from the $(k-1)$th layer as in 
\begin{equation}
\vec{s}(p_j,g_i) = \text{LeakyReLU}\Big( \big( \mat{W}_N \vec{h}^{k-1}_N (g_i) + \vec{b}_N \big) * \big( \mat{W}_E \vec{h}^{k-1}_E (p_j) +\vec{b}_E \big) \Big)
\label{eq:4attn}
\end{equation}
where $*$ denotes element-wise product, $\mat{W}_N$ and $\vec{b}_N$ ($\mat{W}_E$ and $\vec{b}_E$) are the transformation weights and bias of the nodes (the hyperedges) for the attention ready state. This is motivated by the observation that different nodes (genes) contribute differently to the hyperedges (pathways), thus we need proper attentions across the nodes to up- or down-weight their contributions when aggregating their representations to compute the representation of the hyperedge. Once we have the hyperedge attention over nodes, we calculate the hyperedge's representation in layer $k$ from the nodes' representations in layer $k-1$ (equation~\ref{eq:hkE}) where $\sigma$ is the nonlinearity layer (ReLU in our experiment).
\begin{equation}
\vec{h}^k_E (p_j) = \sigma \left( \textstyle\sum\nolimits_{g_i \in p_j} a_E(p_j,g_i) \;  \vec{h}^{k-1}_N (g_i) \right)
\label{eq:hkE}
\end{equation}
\textbf{Node attention over hyperedges.} For a node $g_i \in \mathcal{N}$, in order to update its hidden representation at layer $k$, we aggregate the information from its incident hyperedges using the following attention mechanism. We first calculate the node attention over hyperedges as in
\begin{equation}
a_N(g_i, p_j) = \text{exp}\left( \vec{c}^T \vec{s}(p_j,g_i) \right) \bigg/ \left(\textstyle\sum\nolimits_{p_{j'} \ni g_i} \text{exp}\left( \vec{c}^T \vec{s}(p_{j'},g_i) \right)\right)
\end{equation}
where $\vec{c}$ is the same trainable context vector as used in hyperedge attention calculation and  the attention ready state $\vec{s}(p_j,g_i)$ for hyperedge-node pair $(p_j,g_i)$ is calculated as in equation~\ref{eq:4attn}. This allows us to have the node's attention over the hyperedges, i.e., we can weight hyperedges' contributions when aggregating their representations to compute the representation of the node. We calculate the node's representation in layer $k$ from the hyperedges' representations in layer $k-1$ as in
\begin{equation}
\vec{h}^k_N (g_i) = \sigma \left( \textstyle\sum \nolimits_{p_j \ni g_i} a_N(g_i, p_j) \;  \vec{h}_E^{k-1} (p_j) \right)
\end{equation}
Note that different from HyperGAT, here  the calculation of hyperedge's and node's attentions share the same underlying dual-attention matrix as shown in Fig.~\ref{fig:SHINE} (b), which is essentially unstandardized covariance matrix. Such parameter sharing across hyperedges and nodes allows us to cross-regulate the learning of their mutual attentions to prevent overfitting. This difference not only allows for simplification of the model, but also is more consistent with the notion of duality. The dual $\mathcal{H}^*$ of the hypergraph $\mathcal{H}$ is a hypergraph with $\mathcal{H}$'s vertices and edges interchanged, and we should have $(\mathcal{H}^*)^* = \mathcal{H}$. It is easily provable that the dual-attentions for $(\mathcal{H}^*)^*$ are the same as those for $\mathcal{H}$. Such a self-dual statement is generally not true for the attentions proposed in HyperGAT due to their unsymmetrical way of calculating the node-level and the edge-level attentions, despite that the HyperGAT attention was termed as ``dual'' attention. For this reason, we term our attention message passing scheme as strongly dual attention message passing. 

\textbf{Hypergraph regularization.}
One important intuition about graph and hypergraph convolutional network is that the learned representations for nodes with similar context of (hyper)edges should be similar. In the case of a simple graph $G=(V,E)$, this is to minimize the summed distance $\sum_{(u,v)\in E}\norm{h_u - h_v}^2$ or its weighted variants. Instead of using it as an explicit regularizer, graph or hypergraph convolutional networks leverage an appropriately defined graph or hypergraph Laplacian. As noted in~\citep{zhou2006learning}, the hypergraph Laplacian is $\mat{\Delta}=\mat{I}-\mat{\Theta}$ where $\mat{I}$ is the identity matrix and $\mat{\Theta}$ is defined as (let $\mat{W}$ be a diagonal matrix with diagonal entries as hyperedge weights)
\begin{gather}
\mat{\Theta} = \mat{D}_v^{-1/2} \mat{H} \mat{W} \mat{D}_e^{-1} \mat{H}^T \mat{D}_v^{-1/2} 
\end{gather}
Here, different from hypergraph convolutional networks, we use explicit regularization on the similarity of representations of nodes with similar hyperedge context. Let $\mat{X}$ be the matrix of the learned nodes' representations, where row $\mat{X}_i=\vec{h}^K_N (g_i)$ and the $K$th layer is the last hypergraph message passing layer. We can define the regularizer as 
\begin{gather}
\mathcal{L}_{reg} = \textstyle\sum\nolimits_{i,j} \left((\mat{X}_i \mat{X}_i^T -2 \mat{X}_i \mat{X}_j^T + \mat{X}_j \mat{X}_j^T )* \mat{\Theta}_{ij} \right)
\end{gather}
Intuitively, the more hyperedges are incident on the node pair $i,j$, the more we should penalize their representational differences. On the other hand, the regularizer down-weights the penalization if a hyperedge connects many nodes or if a node has many incident hyperedges, indicating lack of specificity for hyperedges and nodes, respectively.

\subsection{Weighted Subgraph Attention}
The multiple layers of strongly dual attention message passing allow learning the nodes' and hyperedges' representations. However, the instance for the classification algorithm is a subgraph (e.g., a patient, who has mutations in multiple genes (nodes)). From the hypergraph perspective, a patient $j (1\le j \le n)$ can be considered as a subhypergraph $\mathcal{G}_j$ whose nodes (genes) have mutations in $j$ and are a subset of those of $\mathcal{H}$. This is shown in Fig.~\ref{fig:SHINE} (c) where different node colors in the hypergraph denote different patients. In order to calculate the subgraph's representation from its component nodes' representations at the $K$th layer, we use the following weighted subgraph attention (WSA) mechanism, inspired by~\cite{li2015gated}. In fact, none of the previous hypergraph methods support subgraph inferencing, and we had to add our WSA module to those models for subgraph inferencing as well. We first compute the subgraph attention over nodes (e.g., $g_i$'s) as in 
\begin{equation}
a(\mathcal{G}_j, g_i) = \text{exp}\left( \mat{M}_{ji} \vec{b}^T \vec{h}^K_N(g_i) \right) \bigg/ \left( \textstyle\sum\nolimits_{g_{i'} \in \mathcal{G}_j} \text{exp}\left( \mat{M}_{ji'} \vec{b}^T \vec{h}^K_N(g_{i'}) \right) \right)
\label{eq:sgattn}
\end{equation}
where $\vec{b}$ is a trainable context vector, $\mat{M}$ is the mutation rate feature matrix with each row corresponding to a patient and each column corresponding to a gene. Thus, equation~\ref{eq:sgattn} is a mutation rate weighted subgraph attention mechanism. This choice conforms to the intuition that the rate of a mutation is more informative than a categorical indicator of the mutation's occurrence. With these subgraph level attentions, we compute the patient (subgraph) representation from the $K$th layer's gene representations as in
\begin{equation}
\vec{h}(\mathcal{G}_j) = \sigma \left( \textstyle\sum\nolimits_{g_i \in \mathcal{G}_j} a(\mathcal{G}_j,g_i) \;  \vec{h}^K_N (g_i) \right)
\end{equation}
We then stacked the learned patient representations to form the new patient feature matrix, as in
\begin{equation}
\mat{S} = [\vec{h}(\mathcal{G}_1)^T \; | \; \vec{h}(\mathcal{G}_2)^T \; | \; ... \; | \; \vec{h}(\mathcal{G}_n)^T]^T
\end{equation}
where each row is a patient (subgraph) embedding. 

\subsection{Inductive Classification on Subgraphs}
Let the learned feature matrix be $\mat{S}$ and feed it into a softmax classifier
\begin{equation}
\mat{Z} = \text{softmax}(\mat{W}^{(1)} \; (ReLU \circ FC)^{(2)} (\mat{S}) \: + \: \mat{W}^{(0)})
\end{equation}
where (2) in the superscript indicates two MLP layers ($FC$=Fully Connected layer). The loss function is defined as the cross-entropy error over all subjects in all classes as in
\begin{equation}
\mathcal{L} = -\sum\nolimits_{j \in \mathcal{Y}_D} \sum\nolimits_{f=1}^F \mat{Y}_{jf} \ln \mat{Z}_{jf} + \mathcal{L}_{reg}
\label{eq:closs}
\end{equation}
where $\mathcal{Y}_D$ is the training set of subjects that have labels and $F$ is the dimension of the output labels, which is equal to the number of classes. $\mat{Y}$ is the label indicator matrix. 
Note that the subgraph attention layer allows us to compute any patient's representation, which effectively eliminates the need for access to test set patient features during training, making the model inductive. Existing models such as HyperGCN and HGNN are transductive, we cascade our subgraph attention layer on top of these models and make them inductive to serve as our comparison models.

\section{Experiments}
We conducted experiments on real-world datasets in genetic medicine. Both datasets have more than 20 different classes, indicating significant complexity of the prediction tasks. These datasets are different in nature, e.g., curated from literature vs. obtained directly from high-throughput sequencing, and multi-class vs. multi-class multi-label. Our experiments are motivated by the fact that massive genomic data call for novel methods and present unique technical challenges, in this case inductive subgraph inferencing on hypergraph. 
The summary statistics for each dataset is shown in Table~\ref{tab:data-stats}, and the description of each dataset is as follows. Most pathways have small to medium sizes, see Table~\ref{tab:data-stats} pathway sizes IQR. In fact, even at the 95th percentile, the pathway size is just over 200. On the other hand, we observed that the larger the pathway (hyperedge), the more subgraph it is incident on, and the less attention our model will give it as a discriminative feature. The DisGeNet and the TCGA-MC3 datasets are publicly available\footnote{DisGeNet: https://www.disgenet.org/; TCGA-MC3: https://gdc.cancer.gov/about-data/publications/mc3-2017}, and this study is approved by Northwestern University Institutional Review Board.

\subsection{Disease Type Prediction with DisGeNet Data}
In this experiment, we have used the DisGeNet dataset~\citep{gkw943} that is a collection of mutated genes involved in human diseases compiled from expert curated repositories, GWAS catalogs, animal models and the scientific literature. In the following text, we abuse terminology to use “gene” to really mean "variants in the gene". We model genes as hypergraph nodes and diseases as hyperedges. Each disease is labeled with one or more of 22 MeSH codes, and the task is a multi-class multi-label classification problem. We used 6:2:2 train:validation:test partition, and the split distribution is shown in Appendix. The DisGeNet dataset has 6226 pathways and 9133 genes involved in 8383 diseases.

\subsection{Cancer Type Prediction with NGS Somatic Mutations Data}
In this experiment, we have used the consensus somatic mutations for TCGA subjects produced by the Multi-Center Mutation Calling in Multiple Cancers (MC3) project~\cite{ellrott2018scalable}. 
Aiming to enable robust cross-tumor-type analyses, the MC3 approach applied an ensemble of 7 mutation-calling algorithms and assigned a PASS identifier to a mutation that was called by 2 or more variant callers out of the total 7 callers~\citep{ellrott2018scalable}. The MC3 approach accounted for variance and batch effects introduced by the rapid advancement of DNA extraction, hybridization-capture, and sequencing over time. Following this approach, we restricted our analysis to PASS calls in order to maintain sample sizes and uniformity in mutation calling. Each subject is labeled with one of 25 cancer types, and the task is a multi-class classification problem. We used 6:2:2 train:validation:test partition, stratified by cancer types, and the split distribution is shown in Appendix. The TCGA-MC3 dataset has 6229 pathways and 18059 genes involved in 9012 subjects in total. 

\begin{table}
\centering
\footnotesize
\caption{Real-world hypergraph datasets for subgraph inference used in our work. For hyperedge sizes, shown are medians and interquartile ranges. From the distribution, it is clear that the hyperedge size has a positive skewness.}
\begin{tabular}{l|c|c|c|c|c}   
\toprule
Dataset & \# hypernodes & \# hyperedges & Hyperedge size & \# classes & \# subgraphs  \\
\midrule 
DisGeNet & 9133 & 6226 & 25 (12 - 57) & 22 & 8383  \\ 
TCGA-MC3 & 18059 & 6229 & 33 (15 - 77) & 25 & 9012  \\ 
\bottomrule
\end{tabular}
\label{tab:data-stats}
\end{table}

\subsection{Baselines}
We compared SHINE with the following state-of-the-art baselines, we use validation datasets to tune parameters and hyperparameters, please see Appendix for details.
\begin{itemize}
\item Hypergraph neural networks (\textbf{HGNN})~\citep{feng2019hypergraph} uses clique expansion to transform the hypergraph to graph and  uses Chebyshev approximation to derive a simplified hypergraph convolution operation. 
\item \textbf{HyperGCN}~\citep{yadati2018hypergcn} represents a hyperedge by a selected pairwise simple edge connecting two most unlike nodes, and adds the remaining nodes in the hyperedge as mediators. 
\item \textbf{HyperGAT}~\citep{ding2020more} learns node representations by aggregating information from nodes to edges and vice versa. Different from SHINE, HyperGAT uses alternating attention instead of strictly dual attention and has no regularization on nodes with similar context of hyperedges. 
\item \textbf{AllSetTransformer} and \textbf{AllDeepSets} are two variants (attention-based and MLP-based respectively) of set-based methods (e.g., compositions of two multiset functions that are permutation invariant on their input multisets) for exploiting hyperedges in hypergraphs~\citep{chien2021you}. 
\item \textbf{SubGNN}~\citep{alsentzer2020subgraph} was applied to the hypergraph by viewing the nodes and hyperedges as two types of vertices of a bipartite graph~\citep{wang2022survey}. 
\item The multilayer perceptron (\textbf{MLP}) baseline evaluates how a simple feed-forward neural network with hypergraph regularization and subgraph attention performs, by  replacing dual attention with MLP.
\item Polygenic risk score (\textbf{PRS}) is currently a widely used standard practice in genetic medicine and calculates disease risk from genotype profile using regression~\citep{choi2020tutorial}. 
\item Non-negative matrix factorization (\textbf{NMF}) discovers low-dimensional structure from high-dimensional multi-omic data and enables inference of complex biological processes~\citep{stein2018enter}. 
\item \textbf{XGBoost} is an end-to-end tree boosting system and a state-of-the-art machine learning method~\citep{chen2016xgboost} that frequently achieves the top results on many machine learning challenges. 
\end{itemize}

To assess whether performance changes are due to added information (e.g., pathway information) and/or better utilization of the added information, we run PRS, NMF, and XGBoost in the following three settings:  gene features only,  pathway features only, and  both gene and pathway features. 

\section{Results}
The held-out test set micro-averaged F1 scores (micro-F1) for  our proposed method SHINE and all the other comparison models are in Table~\ref{tab:exp-res}. Comparing all the models, we can see that SHINE clearly outperforms a comprehensive array of state-of-the-art baselines in various configurations, with non-overlapping standard deviation intervals. PRS is indeed a competitive baseline, as can be seen from its close performance compared with XGBoost that frequently topped many machine learning challenges' leaderboards. Previously state-of-the-art hypergraph neural network models (HyperGCN, HGNN, HyperGAT) do not always outperform PRS and XGBoost (e.g., on the TCGA-MC3 dataset). On the other hand, pathway as features could improve performance if used properly, whether alone or jointly with genes. This comparison shows that genetic pathway information is useful to disease type classification, consistent to the intuition that pathways encode molecular functional mechanisms that underlie the disease etiology. However, properly utilizing such information is non-trivial, as evidenced by the difficulty to outperform PRS and XGBoost models by NMF models and hypergraph models including HyperGCN, HGNN and HyperGAT. Given that difficulty, SHINE still attained the best performance on each dataset. Intuitively speaking, HyperGCN and HGNN focus on similarity regularization: hypergraph nodes with similar context of hyperedges should have similar representations. HyperGAT's attention mechanism gears more towards minimizing the classification loss. SHINE, in an attempt to balance the similarity regularization with the end-to-end classification task via strongly dual attention mechanism, achieved better trade-off between the two objectives and effectively integrated the functional pathway's (hyperedge) information with individual gene's (node) information. 

The importance of the strong duality follows naturally from that SHINE outperforms SubGNN (bipartite) with non-overlapping standard deviation intervals and wide separation. In addition, although MLP has been frequently used to approximate a target function, in the setting of large hypergraph (e.g., both hypergraphs have a few thousand-nodes hyperedges), it can still be quite challenging to approximate an ideal target function and explicit dual attention formulation wins out. The results from both AllDeepSets and AllSetTransformer have non-overlapping standard deviation intervals, in fact wide separation, with their counterparts from SHINE. These results echo with our observations that strongly dual attention explores the hypergraph propagation from a different angle than both AllDeepSets and AllSetTransformer, and suggest that effectively combining both angles could be an interesting future direction. Also note that in general, we have some performance drop when moving from the DisGeNet dataset to the TCGA-MC3 dataset, likely due to the fact that the former uses genetic features from curated literatures and the latter is from high throughput sequencing intended for data-driven discovery. Both complex classification tasks ($>$20 classes) are uniquely challenging because diseases may have overlapping disrupted molecular functions (genetic pathways, hyperedges), especially for the TCGA-MC3 experiment that is distinguishing subcategories of similar diseases as they are loosely all cancers. In addition, both tasks exhibit class distributional shift between the train and the test datasets, as shown in Appendix Tables~\ref{tab:dgndata} and~\ref{tab:mc3data}, and have been designed to require inductive inference on subgraphs with highly variable hyperedge sizes. Strong performance of SHINE on these tasks thus suggests that our model can leverage its relational inductive biases for more robust generalization. Ablation studies further confirmed the contributions from each of the key components, including strictly dual attention massage passing, weighted subgraph attention and hypergraph regularization (see Appendix for details).

\begin{table}
\footnotesize
\centering
\caption{Held-out test set micro-F1 on real-world datasets. Standard deviations are provided from runs with 10 random seeds. SHINE significantly outperforms all the state-of-the-art comparison models. PRS: polygenic risk score. NMF: non-negative matrix factorization. Best model in bold.}
\begin{tabular}{l|l|r|r}  
\toprule
\midrule
Model & Feature & DisGeNet Dataset &      TCGA-MC3 Dataset \\
\multicolumn{1}{r|}{\textit{Metrics}} & & Test Micro F1 & Test Micro F1  \\
\hline
PRS & gene   & 0.6303   & 0.4981    \\
PRS & pathway  & 0.6461  & 0.5047 \\
PRS & gene+pathway  & 0.6512  & 0.5042 \\
\hline
XGBoost & gene   & 0.6259 $\pm$ 0.0012   & 0.4927 $\pm$ 0.0058    \\ 
XGBoost & pathway   & 0.6467 $\pm$ 0.0035  & 0.4936 $\pm$ 0.0092 \\ 
XGBoost & gene+pathway  & 0.6486 $\pm$ 0.0036  & 0.5117 $\pm$ 0.0084 \\ 
\hline
NMF & gene   & 0.6167 $\pm$ 0.0040  & 0.4181 $\pm$ 0.0125    \\ 
NMF & pathway   & 0.5867 $\pm$ 0.0039  & 0.4842 $\pm$ 0.0057 \\ 
NMF & gene+pathway   & 0.5847 $\pm$ 0.0045  & 0.4839 $\pm$ 0.0032 \\ 
\hline
SubGNN (bipartite) & gene+pathway    & 0.6137 $\pm$ 0.0097  & 0.4025 $\pm$ 0.0049 \\
HyperGCN & gene+pathway   & 0.6638 $\pm$ 0.0028 & 0.4384 $\pm$ 0.0095 \\ 
HGNN & gene+pathway    & 0.6809 $\pm$ 0.0027   & 0.4504 $\pm$ 0.0042 \\ 
HyperGAT & gene+pathway    & 0.6495 $\pm$ 0.0050  & 0.4721 $\pm$ 0.0032 \\ 
MLP & gene+pathway    & 0.6331 $\pm$ 0.0056  & 0.4249 $\pm$ 0.0165 \\ 
AllDeepSets & gene+pathway    & 0.6309 $\pm$ 0.0147  & 0.4324 $\pm$ 0.0220 \\ 
AllSetTransformer & gene+pathway    & 0.6355 $\pm$ 0.0160  & 0.4904 $\pm$ 0.0158 \\ 
SHINE & gene+pathway    & \bf{0.6955 $\pm$ 0.0034}  & \bf{0.5319 $\pm$ 0.0049} \\ 
\bottomrule
\end{tabular}
\label{tab:exp-res}
\end{table}

\paragraph{Model interpretation.} SHINE simultaneously learns the representations of nodes and hyperedges, which are then used to learn and inductively infer subgraph representations. This brings some interpretation advantages as it allows assessing pathways (hyperedges) correlations and reasoning multiple molecular functions mutually interacting and collectively contributing to the disease onset and progression. In addition, SHINE has built-in measures to prevent or discourage genes belonging to the same functional class (e.g., promoting immune reactions) from having drastically different representations (e.g., opposite directions), a phenomenon that will pose interpretation difficulty to other models that do not employ SHINE's hypergraph regularization. We identify the top pathways that are enriched in different cancers using the attention weights learned for SHINE, as shown in Table~\ref{tab:mc3-pathway}. From the table, we see that many of the listed pathways reflect innate key events in the development of individual or multiple types of cancers, consistent with genetic and medical knowledge from wet lab (e.g., TNF/Stress Related Signaling~\citep{mercogliano2020tumor}). We showcase interpretations for breast cancer and lung cancer here, and refer the reader to the Appendix for full interpretation of Table~\ref{tab:mc3-pathway}. For breast cancer, TNF$\alpha$ is not only closely involved in its onset, progression and in metastasis formation, but also linked to therapy resistance~\citep{mercogliano2020tumor}. Regarding the 4-1BB pathway, studies have suggested HER2/4-1BB bispecific molecule as a candidate of alternative therapeutic strategy to patients in HER2-positive breast cancer~\citep{hinner2019tumor}. VIP/PACAP and their receptors have prominent roles in transactivation of the Epidermal growth factor (EGF) family and growth effects in breast cancer~\citep{moody2016vip}. For lung cancer, the ErbB3 receptor recycling controlled by neuroregulin receptor degradation protein-1 is linked to lung cancer and small inhibitory RNA (siRNA) to ErbB3 shows promise as a therapeutic approach to treatment of lung adenocarcinoma~\citep{sithanandam2008erbb3}. Lung cancer is also modulated by multiple miRNAs interacting with the TFAP2 family~\citep{kolat2019biological}.

\begin{table*}[t]
\footnotesize
\centering
\renewcommand{\arraystretch}{1.2}
\caption{Top enriched genetic pathways associated with different cancer risks. The text color indicates the source database for pathways that MSigDB integrated: \textcolor{red}{BioCarta}, Reactome, \textcolor{blue}{WikiPathways}, \textcolor{orange}{Pathway Interaction Database}, \textcolor{purple}{KEGG}.}
\begin{tabular}{c|c|c|c}
\toprule
BRCA & LUAD & LGG & HNSC \\
\hline
\textcolor{red}{Stress pathway} &  PTK6 stabilizes HIF1$\alpha$  & \textcolor{purple}{Citrate cycle TCA cycle}  &  Apoptotic factor response  \\
\hline
\textcolor{red}{4-1BB pathway} &  \textcolor{red}{ErbB3 pathway} & \textcolor{blue}{Cytosine methylation} &  Programmed cell death \\
\hline
\textcolor{red}{VIP pathway} &  \textcolor{purple}{Hypertrophic} & \textcolor{blue}{TCA cycle and deficiency} &  MECP2 regulates neuronal  \\
& \textcolor{purple}{cardiomyopathy} & \textcolor{blue}{of pyruvate dehydrogenase} & receptors and channels \\
\hline
\textcolor{red}{CD40 pathway} &  Diseases of metabolism  & \textcolor{blue}{Glutathione metabolism} & \textcolor{orange}{FRA pathway} \\
\hline
\textcolor{red}{TOLL pathway} &  TFAP2 regulates growth  &  Digestion of  &  Caspase activation via \\
&  factors transcription & dietary carbohydrate  & extrinsic apoptotic signalling \\
\bottomrule
\end{tabular}
\label{tab:mc3-pathway}
\end{table*}

\section{Discussion, Limitation and Future Work}
In addition to being significantly more accurate and interpretable, SHINE uses inductive subgraph inferencing that works well with minibatch, and scales well to large scale problems, as showcased by real-world experiments. It is known that GNN suffers from over-smoothing when the number of layers increases, as increasingly globally uniform representation of nodes may be developed. On the other hand, attention could limit this phenomenon by limiting to a restricted set of nodes. The effect of hypergraph regularization, while also smoothing, happens on a local scale as part of a direct optimization objective and does not accumulate with increasing number of layers. Such decoupling between attention and local smoothing allows SHINE to better explore the optimization landscape.

Our work has limitations. We assumed that the hyperedges are known in advance. However, in reality, as our domain knowledge increases and evolves, we need to account for unknown hyperedges and, better, simultaneously discover novel hyperedges from data while predicting disease classes. Such a task has important clinical utilities in genetic medicine to discover new genetic pathways that may underlie disease etiology, and will be our future work. Moreover, strongly dual attention explores the hypergraph propagation from a different angle than both AllDeepSets and AllSetTransformer, and effectively combining both angles could be an interesting future direction. Another line of future work is to derive a hypergraph coarsening model on top of SHINE. SHINE currently has flat hypergraph layout and does not learn hierarchical representations of hypergraphs. The emerging technique of spatial transcriptomics can enable discovery of localized and hierarchical gene expression patterns~\citep{zeng2022statistical,rao2021exploring}. A flexible hypergraph coarsening model that can effectively learn hierarchical network structure out of the hypergraphs can shed light on the organizations of the hyperedges (e.g., pathways representing synergistic molecular functions in certain tissue context). 

From the application point of view, detecting tumor subtypes is often interesting, and we expect to extend our method to such detections using multi-modal data when large shared datasets will become available~\citep{kline2022multimodal}. To certain extent, the TCGA labels we used reflect subtypes of organ-specific primary tumors, e.g., LUAD vs. LUSC in lung cancer, KIRC vs. KIRP in kidney cancer. On the other hand, identifying drivers genes and pathways for cancer types and other disease subtypes continue to be biologically important~\citep{bailey2018comprehensive} and will be increasingly fruitful with simultaneously collected deep genetic and phenotypic data on the same patients~\citep{luo2019integrating,ritchie2015methods}. 

The field of genetic medicine encompasses areas of molecular biology and clinical phenotyping to explore new relationships between disease susceptibility and human genetics. Though appearing as a single field, it revolutionizes the practice of medicine in preventing, modifying and treating many diseases such as cardiovascular disease and cancer~\citep{green2020strategic}. We expect SHINE to be a useful tool in the quest of broadly advancing the knowledge on disease susceptibility. In these real-world applications, a subject’s genetic profile may contain individual characterizing information. Thus, this work should never be used in violation of an individual’s privacy, and the necessary steps of IRB review and execution of data user agreement need to be properly completed prior to the study.

\section{Conclusions}

We proposed a novel framework termed SubHypergraph Inductive Neural nEtwork (SHINE) for inductive subgraph inferencing on hypergraphs, designed for jointly optimizing the objectives of end-to-end subgraph classification and similarity regularization for representations of hypergraph nodes with similar context of hyperedges. We showed that SHINE improved the performance (micro-F1) of the learned model for disease type prediction for complex ($>$20 classes) genetic medicine datasets of different characteristics and under different settings (e.g., multi-class and/or multi-label). Genetic pathways directly correspond to molecular mechanisms and functions, which are more informative than individual genes and are represented as hyperedges in SHINE. The novel formulation of disease classification as a subgraph inferencing problem allows a hypergraph neural network to link correlated pathways, i.e., interacting molecular mechanisms, to disease etiology. This leads to better performance with added interpretability. We compared SHINE with a wide array of state-of-the-art (hyper)graph neural networks, XGBoost, NMF, and PRS models with different configurations of genes and pathways as features. SHINE consistently outperformed all state-of-the-art baselines significantly in each of the disease classification and cancer classification tasks. Feature analysis of the learned pathway groups that are automatically identified by SHINE in a data-driven fashion offered significant clinical insights about multiple molecular mechanisms that interact and are associated with disease types and status. 

\begin{ack}
This work was supported in part by NIH grants R01LM013337 and U01TR003528.
\end{ack}

\appendix
\section{Appendix for SHINE: SubHypergraph Inductive Neural nEtwork}
\paragraph{Datasets details.} In this section, we give additional details on the datasets used in this paper. The DisGeNet dataset is a collection of mutated genes involved in human diseases compiled from expert curated repositories, GWAS catalogs, animal models and the scientific literature. Each disease is labeled with one or more of 22 MeSH codes, and the task is a multi-class multi-label classification problem. We used 6:2:2 train:validation:test partition, and the split distribution is shown in Table~\ref{tab:dgndata}. The DisGeNet dataset has 6226 pathways and 9133 genes involved in 8383 diseases in total. The TCGA-MC3 dataset records somatic mutations for subjects in The Cancer Genome Atlas (TCGA). The genetic variants are stored in a specially formatted file. A row in the file specifies a particular variant (e.g., Single Nucleotide Polymorphism or insertion/deletion), its chromosomal location, and what proportion of the sequencing reads covering that chromosomal location have that variant, among other characteristics. Each subject is labeled with one or more of 25 cancer types, and the task is a multi-class classification problem. We used 6:2:2 train:validation:test partition, stratified by cancer types, and the split distribution is shown in Table~\ref{tab:mc3data}. The TCGA-MC3 dataset has 6229 pathways and 18059 genes involved in 9012 subjects in total.

\paragraph{Genetic pathways.} Genetic pathways are a valuable tool to assist in representing, understanding, and analyzing the complex interactions between molecular functions. The pathways contain multiple genes (can be modeled using hyperedges) and correspond to genetic functions, including regulations, genetic signaling, and metabolic interactions. They have a wide range of applications, including predicting cellular activity and inferring disease types and status~\citep{alon2006introduction}. For a simplified and illustrative example, a signaling pathway p1 (having 20 genes) sensing the environment may govern (the governing function embodied as a pathway p2 having 15 genes) the expression of transcription factors in another signaling pathway p3 (having 23 genes), which then controls (the controlling function embodied as a pathway p4 having 34 genes) the expression of proteins that play roles as enzymes in a metabolic pathway p5 (having 57 genes). In general, there will be partial overlap between pathways p1 and p2, p2 and p3, p3 and p4, p4 and p5, and other potential partial overlaps corresponding to partial overlaps between their corresponding hyperedges.

\begin{table}[!b]
\centering
\caption{Statistics of DisGeNet experiment data. The table includes the distribution of the 22 MeSH categories with more than 100 diseases. The dataset is split into a training set, a validation set and a test set according to a 6:2:2 ratio.}
\begin{tabular}{lrrrrr}
\toprule
MeSH & Description & Total & Train & Val & Test \\
\midrule
C01 & Infections & 221 & 135 & 45 & 41 \\
C04 & Neoplasms & 1010 & 626 & 190 & 194 \\
C05 & Musculoskeletal Diseases & 1266 & 765 & 239 & 262 \\
C06 & Digestive System Diseases & 430 & 238 & 91 & 101 \\
C07 & Stomatognathic Diseases & 242 & 156 & 50 & 36 \\
C08 & Respiratory Tract Diseases & 235 & 137 & 52 & 46 \\
C09 & Otorhinolaryngologic Diseases & 299 & 188 & 55 & 56 \\
C10 & Nervous System Diseases & 2960 & 1769 & 619 & 572 \\
C11 & Eye Diseases & 756 & 470 & 150 & 136 \\
C12 & Male Urogenital Diseases & 537 & 337 & 102 & 98 \\
C13 & Female Urogenital Diseases and & 640 & 402 & 118 & 120 \\
 & Pregnancy Complications & & & & \\
C14 & Cardiovascular Diseases & 746 & 441 & 147 & 158 \\
C15 & Hemic and Lymphatic Diseases & 624 & 392 & 108 & 124 \\
C16 & Congenital, Hereditary, and Neonatal & 3648 & 2168 & 725 & 755 \\
 & Diseases and Abnormalities  & & & & \\
C17 & Skin and Connective Tissue Diseases & 789 & 459 & 142 & 188 \\
C18 & Nutritional and Metabolic Diseases & 1277 & 725 & 271 & 281 \\
C19 & Endocrine System Diseases & 535 & 327 & 107 & 101 \\
C20 & Immune System Diseases & 415 & 249 & 87 & 79 \\
C23 & Pathological Conditions, Signs and Symptoms & 1795 & 1065 & 387 & 343 \\
C25 & Chemically-Induced Disorders & 135 & 80 & 29 & 26 \\
F01 & Behavior and Behavior Mechanisms & 267 & 164 & 62 & 41 \\
F03 & Mental Disorders & 501 & 295 & 123 & 83 \\
\bottomrule
\end{tabular}
\label{tab:dgndata}
\end{table}

\begin{table}
\centering
\caption{Statistics of TCGA-MC3 experiment data. The table includes the distribution of the 25 cancer types with more than 100 subjects. The dataset is split into a training set, a validation set and a test set according to a 6:2:2 ratio.}
\begin{tabular}{lrrrrr}
\toprule
Cancer & Description & Total & Train & Val & Test \\
\midrule
BLCA & Bladder Urothelial Carcinoma & 411 & 247 & 82 & 82 \\
BRCA & Breast invasive carcinoma & 791 & 475 & 158 & 158 \\
CESC & Cervical squamous cell carcinoma & 289 & 173 & 58 & 58 \\
 & and endocervical adenocarcinoma  & & & & \\
COAD & Colon adenocarcinoma & 288 & 173 & 57 & 58 \\
ESCA & Esophageal carcinoma & 184 & 110 & 37 & 37 \\
GBM & Glioblastoma multiforme & 309 & 185 & 62 & 62 \\
HNSC & Head and Neck squamous cell carcinoma & 507 & 304 & 102 & 101 \\
KIRC & Kidney renal clear cell carcinoma & 368 & 220 & 74 & 74 \\
KIRP & Kidney renal papillary cell carcinoma & 281 & 169 & 56 & 56 \\
LAML & Acute Myeloid Leukemia & 137 & 83 & 27 & 27 \\
LGG & Brain Lower Grade Glioma & 510 & 306 & 102 & 102 \\
LIHC & Liver hepatocellular carcinoma & 363 & 217 & 73 & 73 \\
LUAD & Lung adenocarcinoma & 512 & 307 & 103 & 102 \\
LUSC & 	Lung squamous cell carcinoma & 480 & 288 & 96 & 96 \\
OV & 	Ovarian serous cystadenocarcinoma & 409 & 245 & 82 & 82 \\
PAAD & Pancreatic adenocarcinoma  & 175 & 105 & 35 & 35 \\
PCPG & Pheochromocytoma and Paraganglioma & 178 & 107 & 35 & 36 \\
PRAD & 	Prostate adenocarcinoma & 493 & 295 & 99 & 99 \\
SARC & Sarcoma & 236 & 142 & 47 & 47 \\
SKCM & Skin Cutaneous Melanoma & 466 & 280 & 93 & 93 \\
STAD & Stomach adenocarcinoma & 438 & 262 & 88 & 88 \\
TGCT & Testicular Germ Cell Tumors & 128 & 77 & 25 & 26 \\
THCA & Thyroid carcinoma & 490 & 294 & 98 & 98 \\
THYM & Thymoma & 122 & 74 & 24 & 24 \\
UCEC & Uterine Corpus Endometrial Carcinoma & 447 & 268 & 90 & 89 \\
\bottomrule
\end{tabular}
\label{tab:mc3data}
\end{table}

\paragraph{Genetic variant calling and filtering for TCGA-MC3 dataset.}  
The variants are usually of high dimensionality. For example, in the TCGA-MC3 dataset, even after we retain only the variants that received PASS identifiers, there are still around 3 million variants. Thus, we choose to aggregate their counts according to the affected genes to avoid impractically large matrices. We aggregate genetic variant count at gene level and sum up all the alternative allele counts and reference allele counts in a gene. We calculate the mutation rate for a gene $g$ as in equation~\ref{eq:gcnt}, 
\begin{equation}
\mu(g) = \frac{\sum_{v \in g} C_{ALT}(v)} { \sum_{v \in g} C_{ALT}(v) + \sum_{v \in g} C_{REF}(v)}
\label{eq:gcnt}
\end{equation}
where variant $v$ belongs to the gene $g$, $C_{ALT}(v)$ is the read depth supporting the variant (alternative) allele in tumor sequencing data and $C_{REF}(v)$ is the read depth supporting the reference allele (non-mutated) in tumor sequencing data.

\paragraph{Parameter and hyperparameter tuning for models.} For SHINE and other hypergraph methods, the hyperparameter of hidden dimension $d$ is tuned on the validation dataset with choices from 100 to 1000, at increments of 100. Deep neural network models are often randomly initialized, thus we also run initialization 10 times and report the averages and standard deviations. For the comparison hypergraph neural network models, we used the implementations by the original authors. The hyperparameters were tuned on the validation set using choice grids according to respective papers, or when unspecified, from default grids as with our proposed method (learning rate $\in [0.001, 0.002, 0.005]$, weight decay $\in [0.0001, 0.0005]$, dropout rate $\in [0.4, 0.5, 0.6]$). For PRS, the regularization coefficient $C$ is tuned on the validation dataset with choices from a geometric sequence from 0.001 to 1000 at a multiplying ratio of 10. For NMF, the number of factors is tuned on the validation dataset with choices from 100 to 1000, at increments of 100. For XGBoost, we tuned max tree depth (3, 5, 10), the number of estimators (from 100 to 1000, at increments of 100), and min child weight (0.01, 0.1, 1, 10, 100), using the validation set. For models requiring random initialization, we run initializations 10 times with different seeds and report the averages and standard deviations. We varied the number K of layers from 1 to 4, and found that 2 layers to give the best results for SHINE. 

Regarding sensitivity to the hidden dimension, in general, the performance is less sensitive to the hidden dimensions when it is sufficiently big ($\ge$300), with <0.05 change in micro-F1 score. Smaller hidden dimensions (100-200) can lead to >0.05 micro-F1 drop, likely due to insufficient representation power. The optimal hidden dimension is 300 for the TCGA-MC3 dataset and 600 for the DisGeNet dataset. The performance also shows <0.05 change in micro-F1 score when varying other hyperparameters including learning rate, weight decay, dropout rate in their respective grids as specified above. 

\paragraph{Computational complexity.} The complexity of SHINE scales as the following factors grow: the numbers of layers and nodes, the number and size of hyperedges, the size of hidden dimensions, and finally the number and size of subhypergraphs. We implement SHINE  on PyTorch, and run it on NVIDIA V100 GPUs. We train SHINE for up to 6000 epochs using Adam~\citep{kinga2015method} and stop training if the validation loss does not decrease for 10 consecutive epochs. The TCGA-MC3 dataset's training times are: MLP $\sim$5 min, HyperGCN $\sim$7 min, AllSetTransformer $\sim$20 min, AllDeepSet $\sim$20 min, SHINE $\sim$30 min, HGNN $\sim$30 min, HyperGAT $\sim$30 min, SubGNN >1 day (excluding prebuild time). The DisGeNet dataset's training times are: MLP $\sim$5 min, HyperGCN $\sim$6 min, AllDeepSet $\sim$13 min, HGNN $\sim$15 min, HyperGAT $\sim$15 min, AllSetTransformer $\sim$16 min, SHINE $\sim$20 min, SubGNN >1 day (excluding prebuild time).

\paragraph{Ablation study.} To investigate the contribution of key components (e.g., the strictly dual attention massage passing, the usage of hypergraph regularization) in the proposed algorithm to the overall method, we performed an ablation analysis. The previous state-of-the-art hypergraph neural network models in fact serve as some of the steps in the ablation. For example, HyperGAT does not have strict dual attention message passing and does not employ hypergraph regularization. HGNN and HyperGCN apply hypergraph convolution instead of attention message passing. HyperGCN, compared to HGNN, applies approximate hypergraph convolution by representing a hyperedge by a selected pairwise simple edge connecting two most unlike nodes, and adding the remaining nodes in the hyperedge as mediators. To evaluate the efficacy of the weighted subgraph attention (WSA), we consider a subgraph simply the sum of the nodes (genes) that are of interest (with mutations) for each patient (subgraph). Finally, we added SHINE with no hypergraph regularization to evaluate the regularization effectiveness. The ablation analysis results are shown in Table~\ref{tab:ablation-res}. From the results, it is clear that SHINE's strictly dual attention message passing outperforms HyperGAT without strictly dual attention message passing. We can see that adding hypergraph regularization further improves performance, in fact, with improvement beyond standard deviation intervals of the regularization-ablated model on both datasets. The weighted subgraph attention (WSA) ablation leads to a larger performance drop than hypergraph regularization ablation, which corroborates the importance of the WSA step. We also notice that the performance drop due to WSA ablation on the TCGA-MC3 dataset is larger than that on the DisGeNet dataset. This is consistent with the fact that the TCGA-MC3 dataset has denser hypergraph and larger subgraphs than the DisGeNet dataset. This is also consistent with the fact that differentiating among cancer subtypes is a more complex and nuanced task than differentiating among disease categories. These observations collectively argue for the benefits of weighted subgraph attention over direct aggregation such as sum, and more increasingly so for larger datasets and more complex tasks.

\begin{table}
\centering
\caption{Ablation Analysis: Held-out test set micro-F1 on real-world datasets. Standard deviations are provided from runs with 10 random seeds. SHINE significantly outperforms all the state-of-the-art comparison models. Best model in bold.}
\begin{tabular}{l|r|r}  
\toprule
\midrule
Model & DisGeNet Dataset &      TCGA-MC3 Dataset \\
\multicolumn{1}{r|}{\textit{Metrics}} & Test Micro F1 & Test Micro F1  \\
\hline
HyperGCN (approx. hypergraph convolution)   & 0.6638 $\pm$ 0.0028 & 0.4384 $\pm$ 0.0095 \\ 
HGNN (hypergraph convolution)    & 0.6809 $\pm$ 0.0027   & 0.4504 $\pm$ 0.0042 \\ 
HyperGAT (not strictly dual attention)   & 0.6495 $\pm$ 0.0050  & 0.4721 $\pm$ 0.0032 \\ 
SHINE without weighted subgraph attention   & 0.6472 $\pm$ 0.0053  & 0.4388 $\pm$ 0.0091 \\ 
SHINE without hypergraph regularization   & 0.6829 $\pm$ 0.0059  & 0.5247 $\pm$ 0.0048 \\ 
SHINE   & \bf{0.6955 $\pm$ 0.0034}  & \bf{0.5319 $\pm$ 0.0049} \\ 
\bottomrule
\end{tabular}
\label{tab:ablation-res}
\end{table}

\paragraph{Model interpretation.} SHINE simultaneously learns the representations of nodes and hyperedges, which are then used to learn and inductively infer subgraph representations. This brings model interpretation advantages as it allows assessing pathways (hyperedges) correlations and reasoning multiple molecular functions mutually interacting and collectively contributing to the disease onset. We identify the top pathways that are enriched in different cancers using the attention weights learned for SHINE, as shown in Table 4. From the table, we see that many of the listed pathways reflect innate key events in the development of individual or multiple types of cancers, consistent with genetic and medical knowledge from wet lab (e.g., TNF/Stress Related Signaling~\citep{mercogliano2020tumor}). 

For breast cancer, TNF$\alpha$ is not only closely involved in its onset, progression and in metastasis formation, but also linked to therapy resistance~\citep{mercogliano2020tumor}. Regarding the 4-1BB pathway, studies have suggested HER2/4-1BB bispecific molecule as a candidate of alternative therapeutic strategy to patients in HER2-positive breast cancer~\citep{hinner2019tumor}. VIP/PACAP and their receptors have prominent roles in transactivation of the Epidermal growth factor (EGF) family and growth effects in breast cancer~\citep{moody2016vip}. For lung cancer, the ErbB3 receptor recycling controlled by neuroregulin receptor degradation protein-1 is linked to lung cancer and small inhibitory RNA (siRNA) to ErbB3 shows promise as a therapeutic approach to treatment of lung adenocarcinoma~\citep{sithanandam2008erbb3}. Lung cancer is also modulated by multiple miRNAs interacting with the TFAP2 family~\citep{kolat2019biological}. For lower-grade gliomas, recent studies have reported the association between DNA demethylation and their malignant progressions~\citep{nomura2019dna}. Emerging evidence has also linked the citric acid (TCA) cycle for energy production to fuel the development of certain cancer types, especially those with deregulated oncogene and tumor suppressor expression~\citep{anderson2018emerging}. For head and neck cancer, studies have reported a high percentage of cases with MECP2 copy-number gain and in combination with RAS mutation or amplification~\citep{neupane2016mecp2}. The apoptotic signaling and response pathways involving the mitochondrial pro-apoptotic protein SMAC/Diablo have also been suggested to regulate lipid synthesis that is essential for cancer growth and development~\citep{paul2018new}. 

Of note, the pathways listed in Table 4 for each cancer type play roles in different phases of cancer onset, growth or metastasis, and likely function together in tumorigenesis and progression, as discovered by SHINE. These analyses suggest that besides providing useful and discriminative features, SHINE integrates gene and pathway data to provide insights into functional and molecular mechanisms by linking together multiple pathways that may function together and contribute to cancer development and progression. 

\paragraph{Relevance and impact.} The techniques and results presented in the paper could apply to many diseases through informing genetic medicine practice. In these real-world applications, a subject’s genetic profile may contain individual characterizing information. Thus, this work, or derivatives of it, should never be used in violation of an individual’s privacy. For using individual level dataset such as the TCGA-MC3, the proper steps of IRB review of study and execution of data user agreement need to be properly completed prior to the study, such as done by this study. 

It is important for the machine learning (ML) community to continue being informed about the problems arising in critical application domains such as healthcare and biomedicine that can guide model design. More specifically, explicitly treating hyperedges as first class citizens in the GNN modelling is important, since in this way hyperedges can be the subjects of notions of regularization or attention. This article demonstrated the feasibility to address those needs with our practical considerations of design and implementation choices by  SHINE to advance modern genetic medicine study. We have demonstrated successful applications of SHINE on large-scale genetic medicine datasets, including the TCGA-MC3 dataset that is one of the largest NIH dbGaP datasets. Genetic medicine revolutionizes the practice of medicine in preventing, modifying and treating many diseases such as cardiovascular disease and cancer. In the future, as even larger genetic datasets will be collected through NIH programs such as All of Us and TopMed, we expect SHINE to be a useful tool in the quest of broadly advancing the knowledge on disease susceptibility.

\bibliographystyle{unsrtnat}
\bibliography{shine}

\begin{thebibliography}{62}
\providecommand{\natexlab}[1]{#1}
\providecommand{\url}[1]{\texttt{#1}}
\expandafter\ifx\csname urlstyle\endcsname\relax
  \providecommand{\doi}[1]{doi: #1}\else
  \providecommand{\doi}{doi: \begingroup \urlstyle{rm}\Url}\fi

\bibitem[Liberzon et~al.(2015)Liberzon, Birger, Thorvaldsd{\'o}ttir, Ghandi,
  Mesirov, and Tamayo]{liberzon2015molecular}
Arthur Liberzon, Chet Birger, Helga Thorvaldsd{\'o}ttir, Mahmoud Ghandi, Jill~P
  Mesirov, and Pablo Tamayo.
\newblock The molecular signatures database hallmark gene set collection.
\newblock \emph{Cell systems}, 1\penalty0 (6):\penalty0 417--425, 2015.

\bibitem[Luo and Mao(2021)]{luo2021panther}
Yuan Luo and Chengsheng Mao.
\newblock Panther: pathway augmented nonnegative tensor factorization for
  higher-order feature learning.
\newblock In \emph{Proceedings of the AAAI conference on artificial
  intelligence}, volume~35, pages 371--380, 2021.

\bibitem[Feng et~al.(2019)Feng, You, Zhang, Ji, and Gao]{feng2019hypergraph}
Yifan Feng, Haoxuan You, Zizhao Zhang, Rongrong Ji, and Yue Gao.
\newblock Hypergraph neural networks.
\newblock In \emph{Proceedings of the AAAI Conference on Artificial
  Intelligence}, volume~33, pages 3558--3565, 2019.

\bibitem[Yadati et~al.(2018)Yadati, Nimishakavi, Yadav, Nitin, Louis, and
  Talukdar]{yadati2018hypergcn}
Naganand Yadati, Madhav Nimishakavi, Prateek Yadav, Vikram Nitin, Anand Louis,
  and Partha Talukdar.
\newblock Hypergcn: A new method of training graph convolutional networks on
  hypergraphs.
\newblock \emph{arXiv preprint arXiv:1809.02589}, 2018.

\bibitem[Yadati(2020)]{yadati2020neural}
Naganand Yadati.
\newblock Neural message passing for multi-relational ordered and recursive
  hypergraphs.
\newblock \emph{Advances in Neural Information Processing Systems}, 33, 2020.

\bibitem[Zhang et~al.(2020)Zhang, Zou, and Ma]{zhang2019hyper}
Ruochi Zhang, Yuesong Zou, and Jian Ma.
\newblock Hyper-sagnn: a self-attention based graph neural network for
  hypergraphs.
\newblock In \emph{International Conference on Learning Representations}, 2020.

\bibitem[Ding et~al.(2020)Ding, Wang, Li, Li, and Liu]{ding2020more}
Kaize Ding, Jianling Wang, Jundong Li, Dingcheng Li, and Huan Liu.
\newblock Be more with less: Hypergraph attention networks for inductive text
  classification.
\newblock \emph{arXiv preprint arXiv:2011.00387}, 2020.

\bibitem[Meng et~al.(2018)Meng, Mouli, Ribeiro, and Neville]{meng2018subgraph}
Changping Meng, S~Chandra Mouli, Bruno Ribeiro, and Jennifer Neville.
\newblock Subgraph pattern neural networks for high-order graph evolution
  prediction.
\newblock In \emph{Proceedings of the AAAI Conference on Artificial
  Intelligence}, volume~32, 2018.

\bibitem[Alsentzer et~al.(2020)Alsentzer, Finlayson, Li, and
  Zitnik]{alsentzer2020subgraph}
Emily Alsentzer, Samuel~G Finlayson, Michelle~M Li, and Marinka Zitnik.
\newblock Subgraph neural networks.
\newblock \emph{arXiv preprint arXiv:2006.10538}, 2020.

\bibitem[Wei et~al.(2020)Wei, Yu, and Sun]{wei2020view}
Xin Wei, Ruixuan Yu, and Jian Sun.
\newblock View-gcn: View-based graph convolutional network for 3d shape
  analysis.
\newblock In \emph{Proceedings of the IEEE/CVF Conference on Computer Vision
  and Pattern Recognition}, pages 1850--1859, 2020.

\bibitem[Mao et~al.(2022)Mao, Yao, and Luo]{mao2022imagegcn}
Chengsheng Mao, Liang Yao, and Yuan Luo.
\newblock Imagegcn: Multi-relational image graph convolutional networks for
  disease identification with chest x-rays.
\newblock \emph{IEEE Transactions on Medical Imaging}, 2022.

\bibitem[Duvenaud et~al.(2015)Duvenaud, Maclaurin, Iparraguirre, Bombarell,
  Hirzel, Aspuru-Guzik, and Adams]{duvenaud2015convolutional}
David~K Duvenaud, Dougal Maclaurin, Jorge Iparraguirre, Rafael Bombarell,
  Timothy Hirzel, Al{\'a}n Aspuru-Guzik, and Ryan~P Adams.
\newblock Convolutional networks on graphs for learning molecular fingerprints.
\newblock In \emph{NeurIPS}, pages 2224--2232, 2015.

\bibitem[Gilmer et~al.(2017)Gilmer, Schoenholz, Riley, Vinyals, and
  Dahl]{gilmer2017neural}
Justin Gilmer, Samuel~S Schoenholz, Patrick~F Riley, Oriol Vinyals, and
  George~E Dahl.
\newblock Neural message passing for quantum chemistry.
\newblock In \emph{ICML}, pages 1263--1272. JMLR. org, 2017.

\bibitem[Yao et~al.(2019)Yao, Mao, and Luo]{yao2019graph}
Liang Yao, Chengsheng Mao, and Yuan Luo.
\newblock Graph convolutional networks for text classification.
\newblock In \emph{AAAI}, 2019.

\bibitem[Peng et~al.(2018)Peng, Li, He, Liu, Bao, Wang, Song, and
  Yang]{peng2018large}
Hao Peng, Jianxin Li, Yu~He, Yaopeng Liu, Mengjiao Bao, Lihong Wang, Yangqiu
  Song, and Qiang Yang.
\newblock Large-scale hierarchical text classification with recursively
  regularized deep graph-cnn.
\newblock In \emph{WWW}, pages 1063--1072, 2018.

\bibitem[Li et~al.(2019)Li, Jin, and Luo]{li2019classifying}
Yifu Li, Ran Jin, and Yuan Luo.
\newblock Classifying relations in clinical narratives using segment graph
  convolutional and recurrent neural networks (seg-gcrns).
\newblock \emph{Journal of the American Medical Informatics Association},
  26\penalty0 (3):\penalty0 262--268, 2019.

\bibitem[Zitnik et~al.(2018)Zitnik, Agrawal, and Leskovec]{zitnik2018modeling}
Marinka Zitnik, Monica Agrawal, and Jure Leskovec.
\newblock Modeling polypharmacy side effects with graph convolutional networks.
\newblock \emph{Bioinformatics}, 34\penalty0 (13):\penalty0 i457--i466, 2018.

\bibitem[Mao et~al.(2019)Mao, Yao, and Luo]{mao2019medgcn}
Chengsheng Mao, Liang Yao, and Yuan Luo.
\newblock Medgcn: Graph convolutional networks for multiple medical tasks.
\newblock \emph{arXiv preprint arXiv:1904.00326}, 2019.

\bibitem[Kipf and Welling(2016)]{kipf2016semi}
Thomas~N Kipf and Max Welling.
\newblock Semi-supervised classification with graph convolutional networks.
\newblock \emph{arXiv preprint arXiv:1609.02907}, 2016.

\bibitem[Vinyals et~al.(2015)Vinyals, Bengio, and Kudlur]{vinyals2015order}
Oriol Vinyals, Samy Bengio, and Manjunath Kudlur.
\newblock Order matters: Sequence to sequence for sets.
\newblock \emph{arXiv preprint arXiv:1511.06391}, 2015.

\bibitem[Ying et~al.(2018)Ying, You, Morris, Ren, Hamilton, and
  Leskovec]{ying2018hierarchical}
Rex Ying, Jiaxuan You, Christopher Morris, Xiang Ren, William~L Hamilton, and
  Jure Leskovec.
\newblock Hierarchical graph representation learning with differentiable
  pooling.
\newblock \emph{arXiv preprint arXiv:1806.08804}, 2018.

\bibitem[Hamilton(2020)]{hamilton2020graph}
William~L Hamilton.
\newblock Graph representation learning.
\newblock \emph{Synthesis Lectures on Artifical Intelligence and Machine
  Learning}, 14\penalty0 (3):\penalty0 1--159, 2020.

\bibitem[Zhou et~al.(2006)Zhou, Huang, and Sch{\"o}lkopf]{zhou2006learning}
Dengyong Zhou, Jiayuan Huang, and Bernhard Sch{\"o}lkopf.
\newblock Learning with hypergraphs: Clustering, classification, and embedding.
\newblock \emph{Advances in neural information processing systems},
  19:\penalty0 1601--1608, 2006.

\bibitem[Jin et~al.(2019)Jin, Cao, Zhang, Sun, Deng, and Ji]{jin2019hypergraph}
Taisong Jin, Liujuan Cao, Baochang Zhang, Xiaoshuai Sun, Cheng Deng, and
  Rongrong Ji.
\newblock Hypergraph induced convolutional manifold networks.
\newblock In \emph{IJCAI}, pages 2670--2676, 2019.

\bibitem[Jiang et~al.(2019)Jiang, Wei, Feng, Cao, and Gao]{jiang2019dynamic}
Jianwen Jiang, Yuxuan Wei, Yifan Feng, Jingxuan Cao, and Yue Gao.
\newblock Dynamic hypergraph neural networks.
\newblock In \emph{IJCAI}, pages 2635--2641, 2019.

\bibitem[Satchidanand et~al.(2015)Satchidanand, Ananthapadmanaban, and
  Ravindran]{satchidanand2015extended}
Sai~Nageswar Satchidanand, Harini Ananthapadmanaban, and Balaraman Ravindran.
\newblock Extended discriminative random walk: A hypergraph approach to
  multi-view multi-relational transductive learning.
\newblock In \emph{IJCAI}, pages 3791--3797, 2015.

\bibitem[Feng et~al.(2018)Feng, He, Liu, Nie, and Chua]{feng2018learning}
Fuli Feng, Xiangnan He, Yiqun Liu, Liqiang Nie, and Tat-Seng Chua.
\newblock Learning on partial-order hypergraphs.
\newblock In \emph{Proceedings of the 2018 World Wide Web Conference}, pages
  1523--1532, 2018.

\bibitem[Tu et~al.(2018)Tu, Cui, Wang, Wang, and Zhu]{tu2018structural}
Ke~Tu, Peng Cui, Xiao Wang, Fei Wang, and Wenwu Zhu.
\newblock Structural deep embedding for hyper-networks.
\newblock In \emph{Proceedings of the AAAI Conference on Artificial
  Intelligence}, volume~32, 2018.

\bibitem[Chien et~al.(2021)Chien, Pan, Peng, and Milenkovic]{chien2021you}
Eli Chien, Chao Pan, Jianhao Peng, and Olgica Milenkovic.
\newblock You are allset: A multiset function framework for hypergraph neural
  networks.
\newblock \emph{arXiv preprint arXiv:2106.13264}, 2021.

\bibitem[Huang and Zitnik(2020)]{huang2020graph}
Kexin Huang and Marinka Zitnik.
\newblock Graph meta learning via local subgraphs.
\newblock \emph{Advances in Neural Information Processing Systems}, 33, 2020.

\bibitem[Sun et~al.(2021)Sun, Li, Peng, Wu, Ning, Yu, and He]{sun2021sugar}
Qingyun Sun, Jianxin Li, Hao Peng, Jia Wu, Yuanxing Ning, Phillip~S Yu, and
  Lifang He.
\newblock Sugar: Subgraph neural network with reinforcement pooling and
  self-supervised mutual information mechanism.
\newblock \emph{arXiv preprint arXiv:2101.08170}, 2021.

\bibitem[Wang et~al.(2019)Wang, Ji, Shi, Wang, Ye, Cui, and
  Yu]{wang2019heterogeneous}
Xiao Wang, Houye Ji, Chuan Shi, Bai Wang, Yanfang Ye, Peng Cui, and Philip~S
  Yu.
\newblock Heterogeneous graph attention network.
\newblock In \emph{The world wide web conference}, pages 2022--2032, 2019.

\bibitem[Zhang et~al.(2019)Zhang, Song, Huang, Swami, and
  Chawla]{zhang2019heterogeneous}
Chuxu Zhang, Dongjin Song, Chao Huang, Ananthram Swami, and Nitesh~V Chawla.
\newblock Heterogeneous graph neural network.
\newblock In \emph{Proceedings of the 25th ACM SIGKDD international conference
  on knowledge discovery \& data mining}, pages 793--803, 2019.

\bibitem[Fu et~al.(2019)Fu, Xiong, Philip, Tao, and Zhu]{fu2019metapath}
Yuwei Fu, Yun Xiong, S~Yu Philip, Tianyi Tao, and Yangyong Zhu.
\newblock Metapath enhanced graph attention encoder for hins representation
  learning.
\newblock In \emph{2019 IEEE International Conference on Big Data (Big Data)},
  pages 1103--1110. IEEE, 2019.

\bibitem[Fu et~al.(2020)Fu, Zhang, Meng, and King]{fu2020magnn}
Xinyu Fu, Jiani Zhang, Ziqiao Meng, and Irwin King.
\newblock Magnn: Metapath aggregated graph neural network for heterogeneous
  graph embedding.
\newblock In \emph{Proceedings of The Web Conference 2020}, pages 2331--2341,
  2020.

\bibitem[Hu et~al.(2020)Hu, Dong, Wang, and Sun]{hu2020heterogeneous}
Ziniu Hu, Yuxiao Dong, Kuansan Wang, and Yizhou Sun.
\newblock Heterogeneous graph transformer.
\newblock In \emph{Proceedings of The Web Conference 2020}, pages 2704--2710,
  2020.

\bibitem[Zhang et~al.(2018)Zhang, Cui, Jiang, and Chen]{zhang2018beyond}
Muhan Zhang, Zhicheng Cui, Shali Jiang, and Yixin Chen.
\newblock Beyond link prediction: Predicting hyperlinks in adjacency space.
\newblock In \emph{Proceedings of the AAAI Conference on Artificial
  Intelligence}, volume~32, 2018.

\bibitem[Li et~al.(2015)Li, Tarlow, Brockschmidt, and Zemel]{li2015gated}
Yujia Li, Daniel Tarlow, Marc Brockschmidt, and Richard Zemel.
\newblock Gated graph sequence neural networks.
\newblock \emph{arXiv preprint arXiv:1511.05493}, 2015.

\bibitem[Piñero et~al.(2016)Piñero, Bravo, Queralt-Rosinach,
  Gutiérrez-Sacristán, Deu-Pons, Centeno, García-García, Sanz, and
  Furlong]{gkw943}
Janet Piñero, Àlex Bravo, Núria Queralt-Rosinach, Alba
  Gutiérrez-Sacristán, Jordi Deu-Pons, Emilio Centeno, Javier
  García-García, Ferran Sanz, and Laura~I. Furlong.
\newblock {DisGeNET: a comprehensive platform integrating information on human
  disease-associated genes and variants}.
\newblock \emph{Nucleic Acids Research}, 45\penalty0 (D1):\penalty0 D833--D839,
  10 2016.
\newblock ISSN 0305-1048.
\newblock \doi{10.1093/nar/gkw943}.
\newblock URL \url{https://doi.org/10.1093/nar/gkw943}.

\bibitem[Ellrott et~al.(2018)Ellrott, Bailey, Saksena, Covington, Kandoth,
  Stewart, Hess, Ma, Chiotti, McLellan, et~al.]{ellrott2018scalable}
Kyle Ellrott, Matthew~H Bailey, Gordon Saksena, Kyle~R Covington, Cyriac
  Kandoth, Chip Stewart, Julian Hess, Singer Ma, Kami~E Chiotti, Michael
  McLellan, et~al.
\newblock Scalable open science approach for mutation calling of tumor exomes
  using multiple genomic pipelines.
\newblock \emph{Cell systems}, 6\penalty0 (3):\penalty0 271--281, 2018.

\bibitem[Wang et~al.(2022)Wang, Bo, Shi, Fan, Ye, and Philip]{wang2022survey}
Xiao Wang, Deyu Bo, Chuan Shi, Shaohua Fan, Yanfang Ye, and S~Yu Philip.
\newblock A survey on heterogeneous graph embedding: methods, techniques,
  applications and sources.
\newblock \emph{IEEE Transactions on Big Data}, 2022.

\bibitem[Choi et~al.(2020)Choi, Mak, and O’Reilly]{choi2020tutorial}
Shing~Wan Choi, Timothy Shin-Heng Mak, and Paul~F O’Reilly.
\newblock Tutorial: a guide to performing polygenic risk score analyses.
\newblock \emph{Nature Protocols}, 15\penalty0 (9):\penalty0 2759--2772, 2020.

\bibitem[Stein-O’Brien et~al.(2018)Stein-O’Brien, Arora, Culhane, Favorov,
  Garmire, Greene, Goff, Li, Ngom, Ochs, et~al.]{stein2018enter}
Genevieve~L Stein-O’Brien, Raman Arora, Aedin~C Culhane, Alexander~V Favorov,
  Lana~X Garmire, Casey~S Greene, Loyal~A Goff, Yifeng Li, Aloune Ngom,
  Michael~F Ochs, et~al.
\newblock Enter the matrix: factorization uncovers knowledge from omics.
\newblock \emph{Trends in Genetics}, 34\penalty0 (10):\penalty0 790--805, 2018.

\bibitem[Chen and Guestrin(2016)]{chen2016xgboost}
Tianqi Chen and Carlos Guestrin.
\newblock Xgboost: A scalable tree boosting system.
\newblock In \emph{Proceedings of the 22nd acm sigkdd international conference
  on knowledge discovery and data mining}, pages 785--794, 2016.

\bibitem[Mercogliano et~al.(2020)Mercogliano, Bruni, Elizalde, and
  Schillaci]{mercogliano2020tumor}
Mar{\'\i}a~Florencia Mercogliano, Sof{\'\i}a Bruni, Patricia~V Elizalde, and
  Roxana Schillaci.
\newblock Tumor necrosis factor $\alpha$ blockade: an opportunity to tackle
  breast cancer.
\newblock \emph{Frontiers in oncology}, 10, 2020.

\bibitem[Hinner et~al.(2019)Hinner, Aiba, Jaquin, Berger, D{\"u}rr, Schlosser,
  Allersdorfer, Wiedenmann, Matschiner, Sch{\"u}ler, et~al.]{hinner2019tumor}
Marlon~J Hinner, Rachida Siham~Bel Aiba, Thomas~J Jaquin, Sven Berger,
  Manuela~Carola D{\"u}rr, Corinna Schlosser, Andrea Allersdorfer, Alexander
  Wiedenmann, Gabriele Matschiner, Julia Sch{\"u}ler, et~al.
\newblock Tumor-localized costimulatory t-cell engagement by the 4-1bb/her2
  bispecific antibody-anticalin fusion prs-343.
\newblock \emph{Clinical Cancer Research}, 25\penalty0 (19):\penalty0
  5878--5889, 2019.

\bibitem[Moody et~al.(2016)Moody, Nuche-Berenguer, and Jensen]{moody2016vip}
Terry~W Moody, Bernardo Nuche-Berenguer, and Robert~T Jensen.
\newblock Vip/pacap, and their receptors and cancer.
\newblock \emph{Current opinion in endocrinology, diabetes, and obesity},
  23\penalty0 (1):\penalty0 38, 2016.

\bibitem[Sithanandam and Anderson(2008)]{sithanandam2008erbb3}
G~Sithanandam and LM~Anderson.
\newblock The erbb3 receptor in cancer and cancer gene therapy.
\newblock \emph{Cancer gene therapy}, 15\penalty0 (7):\penalty0 413--448, 2008.

\bibitem[Ko{\l}at et~al.(2019)Ko{\l}at, Ka{\l}uzi{\'n}ska, Bednarek, and
  P{\l}uciennik]{kolat2019biological}
Damian Ko{\l}at, {\.Z}aneta Ka{\l}uzi{\'n}ska, Andrzej~K Bednarek, and
  El{\.z}bieta P{\l}uciennik.
\newblock The biological characteristics of transcription factors ap-2$\alpha$
  and ap-2$\gamma$ and their importance in various types of cancers.
\newblock \emph{Bioscience reports}, 39\penalty0 (3), 2019.

\bibitem[Zeng et~al.(2022)Zeng, Li, Li, and Luo]{zeng2022statistical}
Zexian Zeng, Yawei Li, Yiming Li, and Yuan Luo.
\newblock Statistical and machine learning methods for spatially resolved
  transcriptomics data analysis.
\newblock \emph{Genome biology}, 23\penalty0 (1):\penalty0 1--23, 2022.

\bibitem[Rao et~al.(2021)Rao, Barkley, Fran{\c{c}}a, and
  Yanai]{rao2021exploring}
Anjali Rao, Dalia Barkley, Gustavo~S Fran{\c{c}}a, and Itai Yanai.
\newblock Exploring tissue architecture using spatial transcriptomics.
\newblock \emph{Nature}, 596\penalty0 (7871):\penalty0 211--220, 2021.

\bibitem[Kline et~al.(2022)Kline, Wang, Li, Dennis, Hutch, Xu, Wang, Cheng, and
  Luo]{kline2022multimodal}
Adrienne Kline, Hanyin Wang, Yikuan Li, Saya Dennis, Meghan Hutch, Zhenxing Xu,
  Fei Wang, Feixiong Cheng, and Yuan Luo.
\newblock Multimodal machine learning in precision health.
\newblock \emph{arXiv preprint arXiv:2204.04777}, 2022.

\bibitem[Bailey et~al.(2018)Bailey, Tokheim, Porta-Pardo, Sengupta, Bertrand,
  Weerasinghe, Colaprico, Wendl, Kim, Reardon, et~al.]{bailey2018comprehensive}
Matthew~H Bailey, Collin Tokheim, Eduard Porta-Pardo, Sohini Sengupta, Denis
  Bertrand, Amila Weerasinghe, Antonio Colaprico, Michael~C Wendl, Jaegil Kim,
  Brendan Reardon, et~al.
\newblock Comprehensive characterization of cancer driver genes and mutations.
\newblock \emph{Cell}, 173\penalty0 (2):\penalty0 371--385, 2018.

\bibitem[Luo et~al.(2019)Luo, Mao, Yang, Wang, Ahmad, Arnett, Irvin, and
  Shah]{luo2019integrating}
Y~Luo, C~Mao, Y~Yang, F~Wang, FS~Ahmad, D~Arnett, MR~Irvin, and SJ~Shah.
\newblock Integrating hypertension phenotype and genotype with hybrid
  non-negative matrix factorization.
\newblock \emph{Bioinformatics (Oxford, England)}, 35\penalty0 (8):\penalty0
  1395--1403, 2019.

\bibitem[Ritchie et~al.(2015)Ritchie, Holzinger, Li, Pendergrass, and
  Kim]{ritchie2015methods}
Marylyn~D Ritchie, Emily~R Holzinger, Ruowang Li, Sarah~A Pendergrass, and
  Dokyoon Kim.
\newblock Methods of integrating data to uncover genotype--phenotype
  interactions.
\newblock \emph{Nature Reviews Genetics}, 16\penalty0 (2):\penalty0 85--97,
  2015.

\bibitem[Green et~al.(2020)Green, Gunter, Biesecker, Di~Francesco, Easter,
  Feingold, Felsenfeld, Kaufman, Ostrander, Pavan, et~al.]{green2020strategic}
Eric~D Green, Chris Gunter, Leslie~G Biesecker, Valentina Di~Francesco, Carla~L
  Easter, Elise~A Feingold, Adam~L Felsenfeld, David~J Kaufman, Elaine~A
  Ostrander, William~J Pavan, et~al.
\newblock Strategic vision for improving human health at the forefront of
  genomics.
\newblock \emph{Nature}, 586\penalty0 (7831):\penalty0 683--692, 2020.

\bibitem[Alon(2006)]{alon2006introduction}
Uri Alon.
\newblock \emph{An introduction to systems biology: design principles of
  biological circuits}.
\newblock Chapman and Hall/CRC, 2006.

\bibitem[Kingma and Ba(2015)]{kinga2015method}
DP~Kingma and JL~Ba.
\newblock Adam: A method for stochastic optimization.
\newblock In \emph{ICLR}, 2015.

\bibitem[Nomura et~al.(2019)Nomura, Saito, Aihara, Nagae, Yamamoto, Tatsuno,
  Ueda, Fukuda, Umeda, Tanaka, et~al.]{nomura2019dna}
Masashi Nomura, Kuniaki Saito, Koki Aihara, Genta Nagae, Shogo Yamamoto, Kenji
  Tatsuno, Hiroki Ueda, Shiro Fukuda, Takayoshi Umeda, Shota Tanaka, et~al.
\newblock Dna demethylation is associated with malignant progression of
  lower-grade gliomas.
\newblock \emph{Scientific reports}, 9\penalty0 (1):\penalty0 1--12, 2019.

\bibitem[Anderson et~al.(2018)Anderson, Mucka, Kern, and
  Feng]{anderson2018emerging}
Nicole~M Anderson, Patrick Mucka, Joseph~G Kern, and Hui Feng.
\newblock The emerging role and targetability of the tca cycle in cancer
  metabolism.
\newblock \emph{Protein \& cell}, 9\penalty0 (2):\penalty0 216--237, 2018.

\bibitem[Neupane et~al.(2016)Neupane, Clark, Landini, Birkbak, Eklund, Lim,
  Culhane, Barry, Schumacher, Beroukhim, et~al.]{neupane2016mecp2}
Manish Neupane, Allison~P Clark, Serena Landini, Nicolai~J Birkbak, Aron~C
  Eklund, Elgene Lim, Aedin~C Culhane, William~T Barry, Steven~E Schumacher,
  Rameen Beroukhim, et~al.
\newblock Mecp2 is a frequently amplified oncogene with a novel epigenetic
  mechanism that mimics the role of activated ras in malignancy.
\newblock \emph{Cancer discovery}, 6\penalty0 (1):\penalty0 45--58, 2016.

\bibitem[Paul et~al.(2018)Paul, Krelin, Arif, Jeger, and
  Shoshan-Barmatz]{paul2018new}
Avijit Paul, Yakov Krelin, Tasleem Arif, Rina Jeger, and Varda Shoshan-Barmatz.
\newblock A new role for the mitochondrial pro-apoptotic protein smac/diablo in
  phospholipid synthesis associated with tumorigenesis.
\newblock \emph{Molecular Therapy}, 26\penalty0 (3):\penalty0 680--694, 2018.

\end{thebibliography}

\end{document}